\pdfoutput=1
\documentclass[letterpaper]{article} 
\usepackage{aaai23}  
\usepackage{times}  
\usepackage{helvet}  
\usepackage{courier}  
\usepackage[hyphens]{url}  
\usepackage{graphicx} 
\usepackage{natbib}  
\usepackage{caption} 
\usepackage{algorithm}
\usepackage{algorithmic}
\usepackage{color}
\usepackage{subfigure}
\usepackage{overpic}
\usepackage{multicol}
\usepackage{multirow}
\usepackage{booktabs}
\usepackage{amsmath,amsfonts}
\usepackage{array}
\newcommand{\chongyi}[1]{{\color{red}(Chongyi: {#1})}} 
\usepackage{newfloat}
\usepackage{listings}
\DeclareCaptionStyle{ruled}{labelfont=normalfont,labelsep=colon,strut=off} 
\floatstyle{ruled}
\newfloat{listing}{tb}{lst}{}
\floatname{listing}{Listing}
\title{Underwater Ranker: Learn Which Is Better and How to Be Better}
\author{
    Chunle Guo\textsuperscript{\rm 1}\equalcontrib,
    Ruiqi Wu\textsuperscript{\rm 1}\equalcontrib,
    Xin Jin\textsuperscript{\rm 1},
    Linghao Han\textsuperscript{\rm 1},
    Zhi Chai\textsuperscript{\rm 2},
    Weidong Zhang\textsuperscript{\rm 3},
    Chongyi Li\textsuperscript{\rm 4}\thanks{Corresponding author.}
}
\affiliations{
    \textsuperscript{\rm 1}TMCC, CS, Nankai University \quad
    \textsuperscript{\rm 2}Hisilicon Technologies Co. Ltd.\\
    \textsuperscript{\rm 3}School of Information Engineering, Henan Institute of Science and Technology\\
    \textsuperscript{\rm 4}S-Lab, Nanyang Technological University\\
    guochunle@nankai.edu.cn,\quad \{wuruiqi, xjin, lhhan\}@mail.nankai.edu.cn\\
    chaizhi2@huawei.com,\quad  zwd\_wd@163.com,\quad  chongyi.li@ntu.edu.sg
    
}

\usepackage{bibentry}

\definecolor{wrq_color}{RGB}{16, 172, 132}

\newcommand{\etal}{\textit{et al.}}
\newcommand{\ie}{\textit{i.e.}}

\begin{document}
\maketitle

\begin{abstract}
%
In this paper, we present a ranking-based underwater image quality assessment (UIQA) method, abbreviated as URanker. 
The URanker is built on the efficient conv-attentional image Transformer.
In terms of underwater images, we specially devise (1) the histogram prior that embeds the color distribution  of an underwater image as histogram token to attend global degradation and  (2) the dynamic cross-scale correspondence to model local degradation.
The final prediction depends on the class tokens from  different scales, which comprehensively considers multi-scale dependencies.
With the margin ranking loss, our URanker can accurately rank the order of  underwater images of the same scene enhanced by different underwater image enhancement (UIE) algorithms according to their visual quality.
To achieve that, we also contribute a dataset, URankerSet, containing sufficient  results enhanced by different UIE algorithms and the corresponding perceptual rankings, to train our URanker.
Apart from the good performance of URanker, we found that a simple U-shape UIE network can obtain promising performance when it is coupled with our pre-trained URanker as additional supervision.
In addition, we also propose a normalization tail that can significantly improve the performance of UIE networks.
Extensive experiments demonstrate the state-of-the-art performance of our method. 
The key designs of our method are discussed. 
Our code and dataset is available at \textit{https://github.com/RQ-Wu/UnderwaterRanker}.
\end{abstract}

\begin{figure}[t]
    \centering
    \subfigure[Learn which is better.]{
        \includegraphics[width=\linewidth]{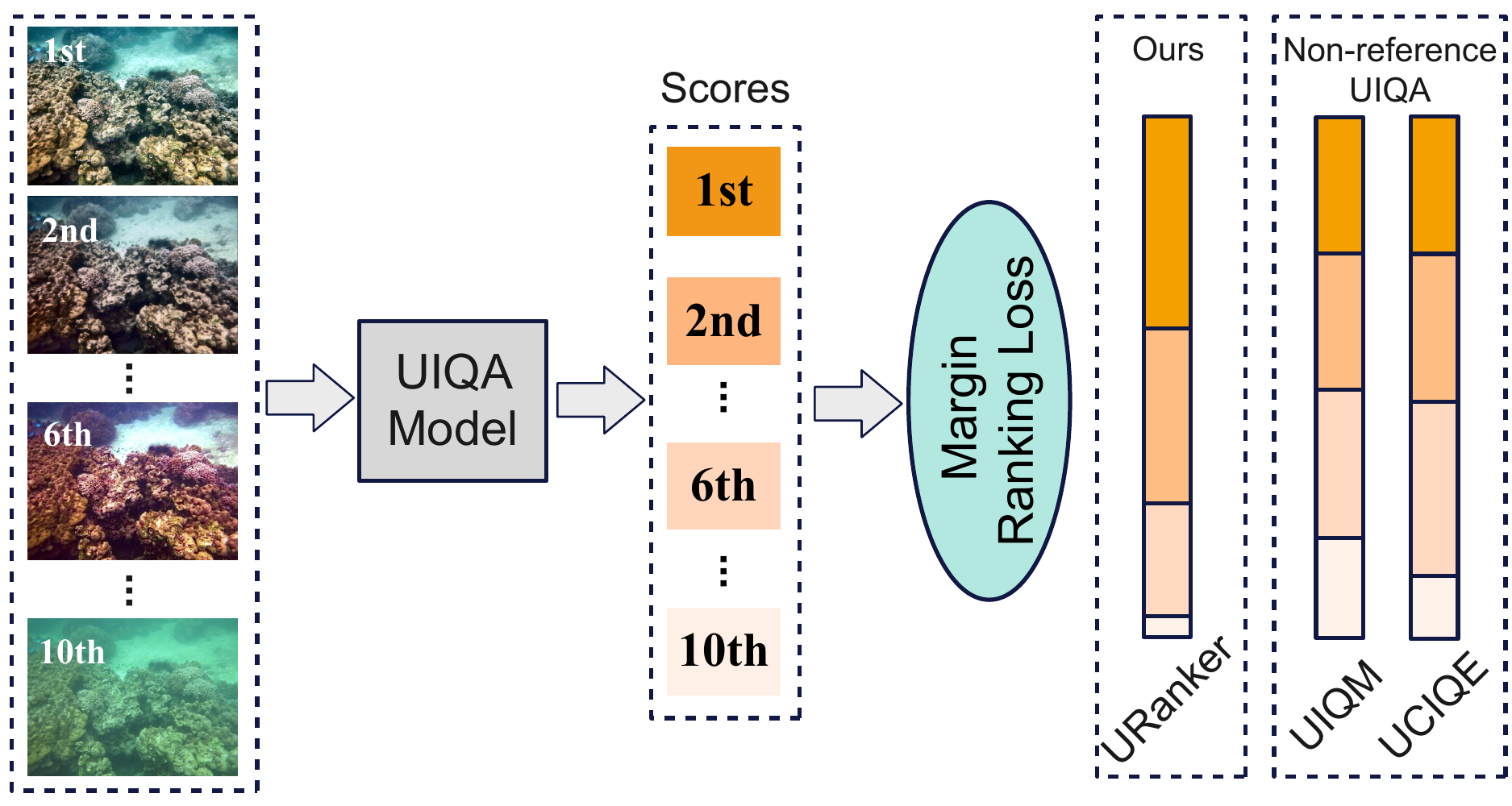}
    }
    \subfigure[Learn how to be better.]{
        \includegraphics[width=\linewidth]{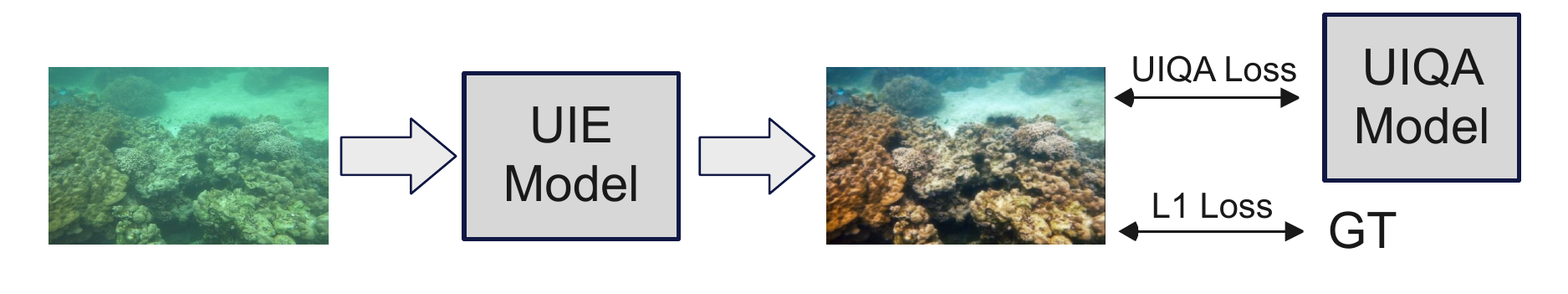}
    }
    \caption{
            (a) illustrates our  URanker that is trained on well-sorted image groups and optimized by margin ranking loss.
            The right part shows the predicted scores of the proposed URanker and two widely-used non-reference UIQA methods, UCIQE and UIQM.
            The ratio of the height of the rectangle represents the ratio of the quality score of the corresponding image, and higher, better. Only our URanker produces the scores  in the decreasing order (align with the ground truth rankings).
            (b) shows the overview of a simple UIE model equipped with our pre-trained URanker as additional supervision which helps a UIE model  obtain better visual quality.
            }
    \label{fig:intro-compare}
\end{figure}

\section{Introduction}
Underwater images commonly suffer from quality degradation issues  such as color cast, low contrast, blurred details, etc.
The wavelength- and distance-dependent light attenuation and scattering are the culprits of these issues~\cite{akkaynak2017space}.
%
To improve the visual quality of underwater images, underwater image enhancement (UIE) techniques have been widely studied.
%
%
With the rapid development of UIE, underwater image quality assessment (UIQA) plays a more and more critical role.
A good UIQA approach not only fairly evaluates the visual quality of underwater images, but also further motivates better UIE algorithms.
%

While existing UIQA approaches show good performance, they still have some limitations.
Although full-reference IQA can objectively evaluate image quality, it is almost impossible to simultaneously obtain both underwater images and the ground truths in situ.
%
%
To bypass this issue, Li \etal~\cite{li2019underwater} proposed a paired underwater image dataset, named UIEB.
UIEB contains real-world underwater images and their reference counterparts.
The reference image is obtained by manually selecting the best enhancement among the  results enhanced by different UIE algorithms.
The advent of UIEB allows one to use full-reference IQA such as PSNR and SSIM to evaluate the quality of underwater images. 
%
However, the provided reference images are not always ideal, leading to inaccurate assessment.
%
There are several non-reference UIQA methods.
%
%
However, they use fixed hand-crafted metrics to evaluate different underwater scenes,  which is inappropriate for underwater images with diverse degradations and captured in different scenes.
%
%
For instance, as shown in Figure~\ref{fig:intro-compare}(a),
existing non-reference UIQA methods like 
UCIQE \cite{yang2015underwater}
and UIQM \cite{panetta2015human}
cannot accurately reflect the quality of underwater images.

To address these problems, we propose a ranking-based UIQA method, abbreviated as URanker.
Instead of designing the hand-crafted metrics or fitting the visual scores, we attempt to rank the order of  different underwater images of the same scene according to their visual quality.
The benefit of using the ranker-learning technique~\cite{liu2017rankiqa,zhang2019ranksrgan} is to avoid the  
interference caused by different degradation and scenes, which are especially obvious in underwater images.
Besides, in contrast to fitting the visual scores, the relative quality order is easier to be learned for an IQA network.
Although there are some ranking-based IQA methods~\cite{liu2017rankiqa, ou2021novel}, they either are originally designed for generic images or ignore the unique characteristics of underwater images.

Different from previous methods, our URanker is built on the efficient conv-attentional image Transformer~\cite{xu2021co}.
Taking the characteristics of underwater images into account, we specially devise two key designs.
Considering that the degrees of color casts significantly influence the visual quality of underwater images, we propose
a histogram prior that embeds the color distribution of an underwater image as histogram token to attend global degradation.
Inspired by the multi-scale features based IQA \cite{ke2021musiq}, we introduce the dynamic cross-scale correspondence to model local degradation.
Besides, we obtain the final prediction by considering multi-scale dependencies according to the class tokens from different scales.
To achieve that, we contribute a dataset, URankerSet, containing sufficient  results enhanced by different UIE algorithms and the corresponding perceptual rankings.
To rank the order of different results, we adopt Bubble Sort with the help of volunteers,
which effectively reflects the human visual perception. 
With the URankerSet, we optimize our URanker using margin ranking loss.
Besides the good performance of URanker for UIQA, since all operations in our URanker are differentiable,
its feedback can be further used as a loss to optimize UIE networks, as illustrated in Figure~\ref{fig:intro-compare}(b).
Further, we propose a simple but effective accessory, normalization tail, for UIE networks, producing significant performance improvements.

Overall, our contributions can be summarized as follows:
\begin{enumerate}
    \item We present a  Transformer network, URanker, for UIQA with novel histogram
prior and dynamic cross-scale correspondence. 
    \item We contribute an underwater image with ranking dataset, URankerSet, that contains underwater images enahnced by different UIE algorithms and the corresponding visual perceptual rankings. 
    \item 
    Apart from the good performance in UIQA, our URanker also improves the performance of UIE networks. The performance can be further improved using our proposed normalization tail.
 
\end{enumerate}


\section{Related Work}

\subsection{Underwater Image Quality Assessment}
Due to the unique characteristics of underwater images, the commonly used IQA metrics~\cite{mittal2012no, talebi2018nima, mittal2012making} 
are not suitable.
Therefore, several IQA specially designed for underwater images have been proposed. 
Yang \etal~\cite{yang2015underwater} proposed the UCIQE, 
which is a linear combination of image chroma, brightness, and saturation.
From the perspective of the human visual system, 
UIQM \cite{panetta2015human} was proposed, which quantifies the colorfulness, sharpness, and contrast of the underwater image, respectively.
UIQM score is calculated by the weighted sum of these three components.
Recently, the Frequency Domain Underwater Metric (FDUM) \cite{yang2021reference} was proposed, which takes the frequency domain features and 
DCP \cite{he2010single} into consideration.
Jiang \etal~\cite{jiang2022underwater} proposed to extract the features of chromatic and luminance
and train an SVM for UIQA.
These UIQA methods achieve good performance in most cases.
Nevertheless, the methods based on hand-craft features~\cite{yang2015underwater,panetta2015human} and statistic regression~\cite{yang2021reference, jiang2022underwater}
do not always hold for diverse underwater images.
More recently, Fu \etal~\cite{fu2022twice} provided a Twice-Mixing framework for UIQA.
It generates a set of mid-quality images by mixing up the original images and the corresponding reference version with different degrees.
Then, a siamese network is adopted to learn their quality rankings.
However, the human visual system is not uniform and 
the mixing-up manner cannot cover the over-enhanced cases.
In our study, we manually sort the  images enhanced by different UIE algorithms,
which generates  more comprehensive results compared to the mixing-up operation.
Besides, we specially design a UIQA method for modeling the global and local degradation of underwater images. Moreover, our method models the multi-scale dependencies via class tokens from different scales.

\subsection{Underwater Image Enhancement}
As a key step for improving the visual quality of underwater images, UIE  has attracted widespread attention.
Existing UIE algorithms can be generally divided into three categories: 
non-physical model-based, physical model-based methods, and data-driven methods.
Non-physical model-based methods~\cite{iqbal2010enhancing, ancuti2012enhancing, fu2017two} 
were proposed to adjust pixel values from the perspective of color balance and contrast.
For physical model-based methods, one line is to 
modify the Dark Channel Prior (DCP)~\cite{he2010single} to make it fit the underwater scenes
\cite{drews2016underwater, li2017hybrid, peng2018generalization}.
Another line is to solve an  underwater image formation model
\cite{galdran2015automatic, li2016underwater, peng2017underwater}.
Recently, with the fast development of deep learning,
many data-driven methods~\cite{li2017watergan, li2020underwater,li2018emerging, li2019underwater, li2021underwater} have been proposed.
The promising results they achieved show their tremendous potential for UIE.
In this study, we demonstrate that a simple UIE network such as U-shape network~\cite{ronneberger2015u} can achieve state-of-the-art performance when it is equipped with our pre-trained URanker as additional supervision and our proposed normalization tail.


\begin{figure*}[!t]
    \centering
    \includegraphics[width=1\textwidth]{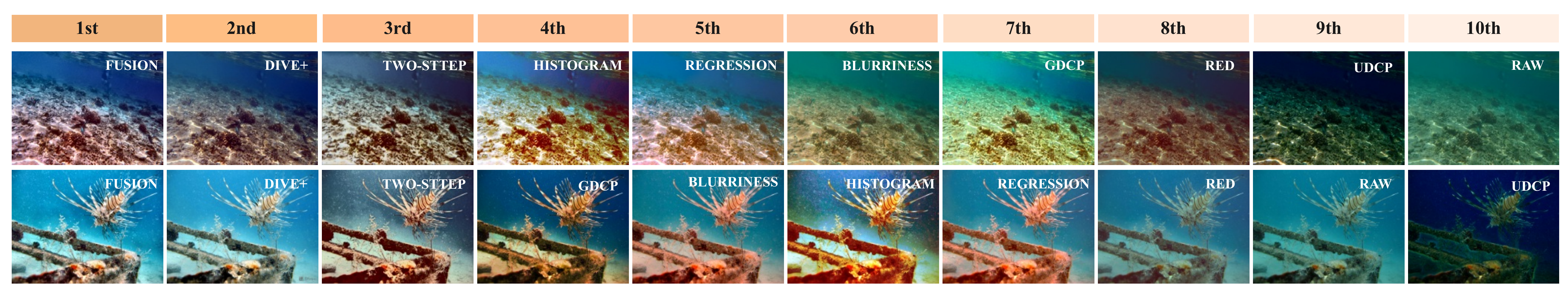}
    \caption{
        Some visual samples of the proposed URankerSet.
        The correspondence ground truth ranking is marked at the top of the figure.
        The UIE algorithm for achieving the enhanced result  is marked at the top-right corner of each image.
    }
    \label{fig:dataset}

\end{figure*}

\begin{figure*}[!t]
    \centering
    \includegraphics[width=.93\textwidth]{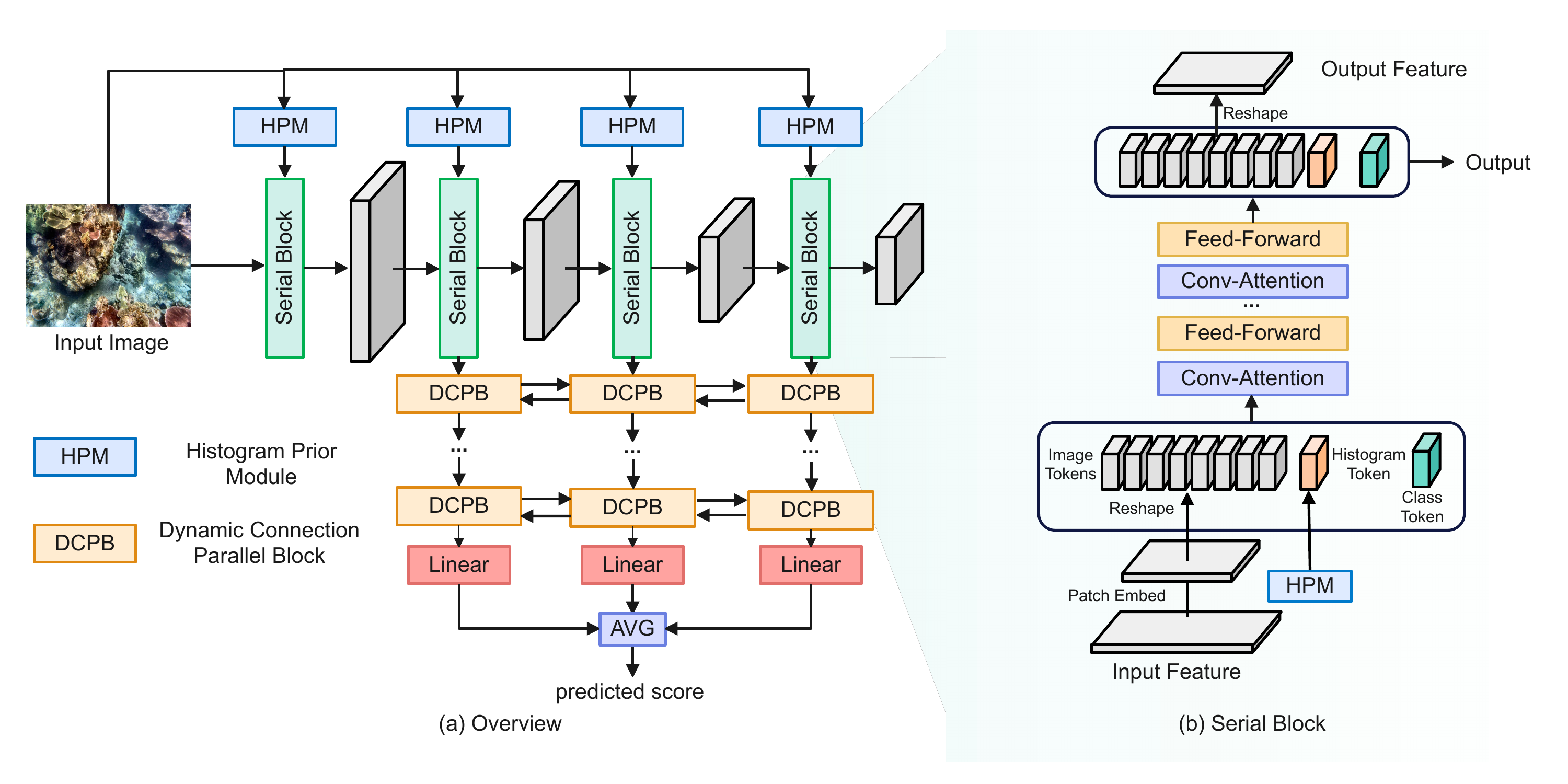}
    \caption{Overview of the proposed URanker.
    The multi-scale features are generated by serial blocks and the
    cross-scale features  are communicated in a parallel fashion.
    In each serial block, the tokenized features are combined with the histogram token from the HPM and the class token, and then processed by the conv-attention and feed-forward layer alternately. The final prediction is the average of the scores from different scales.}
    \label{fig:overview}
\end{figure*}

\section{Proposed URankerSet}

To train the ranking-based UIQA network, we construct an underwater image with rank dataset, called URankerSet.
%
Following the prototype of Li  \etal~\cite{li2019underwater}, we use the same UIE algorithms including 
fusion-based~\cite{ancuti2012enhancing}, two-step-based~\cite{fu2017two}, retinex-based~\cite{fu2014retinex}, UDCP~\cite{drews2016underwater}, regression-based~\cite{li2017hybrid}, GDCP~\cite{peng2018generalization}, Red Channel~\cite{galdran2015automatic}, histogram prior~\cite{li2016underwater}, blurriness-based~\cite{peng2017underwater}, DCP~\cite{he2010single}, MSCNN~\cite{dehaze2}, and dive+\footnote{https://itunes.apple.com/us/app/dive-video-color-correction/ id1251506403?mt=8}
to process 890 raw underwater images provided in  Li \etal's work. 
We discard three of twelve UIE algorithms used in Li \etal' s work as these algorithms produce similar outputs with one of the nine selected UIE algorithms. Otherwise,  similar outputs may affect the accuracy of visual quality ranking.

Specifically, for each raw underwater image and the nine enhanced results, we invite eleven volunteers with image processing experience to perform pairwise comparisons using the monitors with the same model and settings.
We perform Bubble Sort in descending order that is originally designed for arranging a string of numbers in the correct order to rank the visual quality of each set of underwater images (\ie, a raw underwater image and nine  results enhanced by nine different UIE algorithms). 
First, the ten underwater images are randomly arranged from left to right. 
Then, the volunteers are required to compare two adjacent images at a time in left to right order.
In each comparison, the image of the two adjacent images which receive the minority votes is rearranged in descending order from left to right. 
Then, the volunteers continue to cycle through the entire sequence until it completes a pass without switching any images.
In Figure~\ref{fig:dataset}, we show some examples of our URankerSet with visual rankings. 

\section{Proposed URanker}

Recent work has demonstrated that multi-scale features and Transformer structure 
can boost the performance of IQA~\cite{ke2021musiq}.
Inspired by this insight, our URanker is built on the efficient conv-attentional image Transformer~\cite{xu2021co}, which adopts the conv-attention mechanism to replace the multi-head self-attention mechanism for reducing the computational complexity.


The overview of the proposed URanker is illustrated in Figure~\ref{fig:overview}(a). 
Firstly, the input image is fed to a series of serial blocks, in which each serial block  mainly contains a conv-attentional image Transformer, as shown in Figure~\ref{fig:overview}(b). 
In each serial block, the input features are first downsampled 2$\times$ by a patch embedding operation
and are flattened into a group of tokens.
These tokens are concatenated with a learnable class token that represents the features from the current scale and a histogram token produced by the proposed Histogram Prior Module (HPM). 
%
%
%
Then, the conv-attention layer and the feed-forward layer alternately attend all tokens.
After self-attention, the image tokens are reshaped to their original shape and used as input of the next serial block.
In addition, the multi-scale outputs sequentially produced by the serial blocks are sent to the Dynamic Connection Parallel Block (DCPB).
The DCPBs are responsible for integrating cross-scale features in a parallel manner. 
Besides, the class tokens from different scales of the DCPB are sent to the corresponding Linear layer 
for predicting the scores of different scales.
The final prediction is the average of these scores.
In what follows, the key components of the proposed URanker are introduced. More detailed structures and parameters can be found in the supplementary material.

\subsection{Histogram Prior Module}
\label{sec:hpm}
 
\begin{figure}[t]
    \centering
    \includegraphics[width=.92\linewidth]{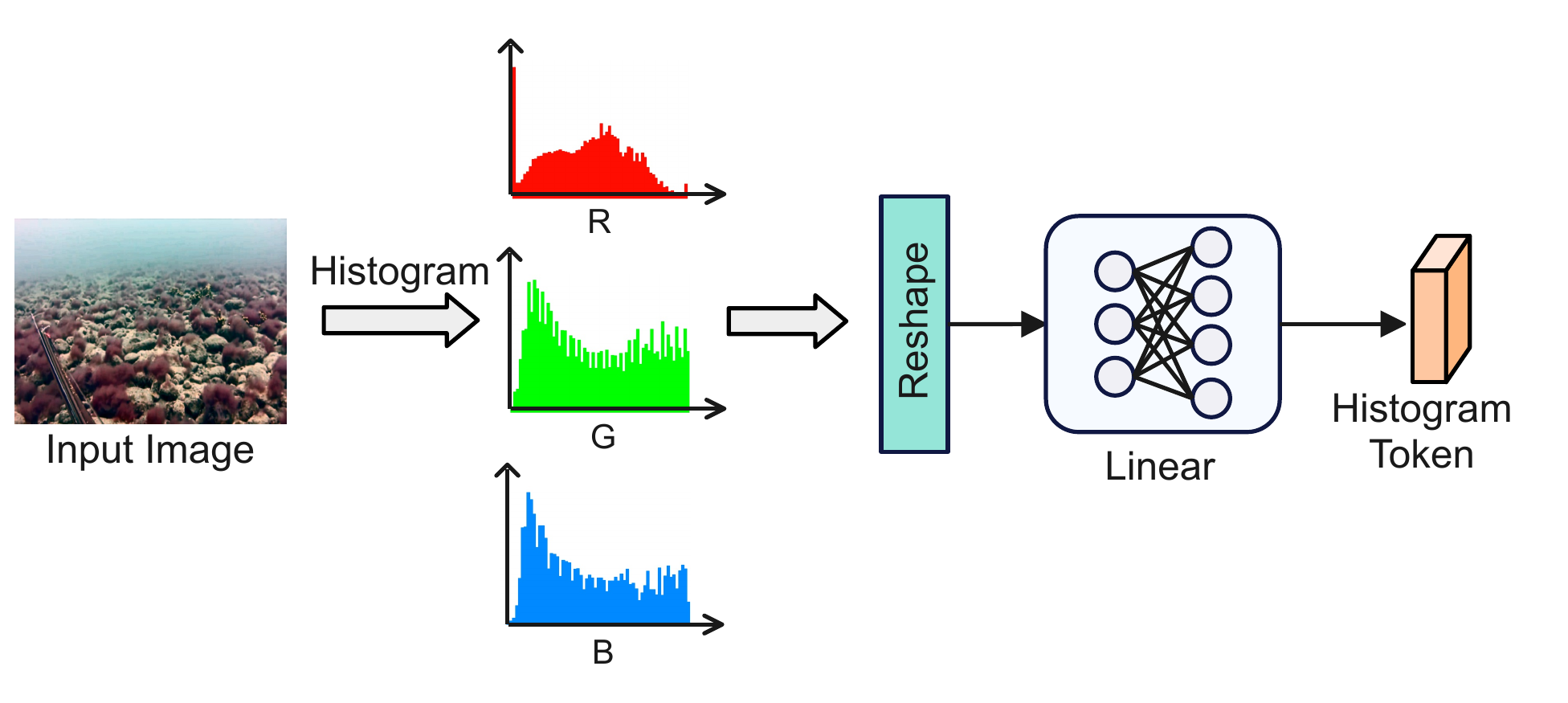}
    \caption{Illustration of our histogram prior module (HPM).
    The input image is first quantified into histograms,
    and the results are reshaped into a vector.
    Then, a Linear layer is followed to generate a histogram token.}
    \label{fig:hpm}
\end{figure}

Global degradation type is essential for assessing the quality of underwater images.
Empirically, the color distribution of high-quality images tends to be uniform.
%
%
Based on this observation, we utilize the color histograms of underwater image as a prior 
and design a simple yet effective module to embed this prior into the proposed URanker.
Figure~\ref{fig:hpm} presents the overview of the proposed HPM. 
Specifically, for an input underwater image, 
we first calculate the histogram of each channel $\mathcal{H} \in {R}^{B \times 3}$,
where $B$ is the number of bins and is set to $64$.
Then $\mathcal{H}$ is flattened into a vector of length $3B$ and
fed to a Linear layer to generate the histogram token.

\begin{figure}[t]
    \begin{overpic}[width=\linewidth]{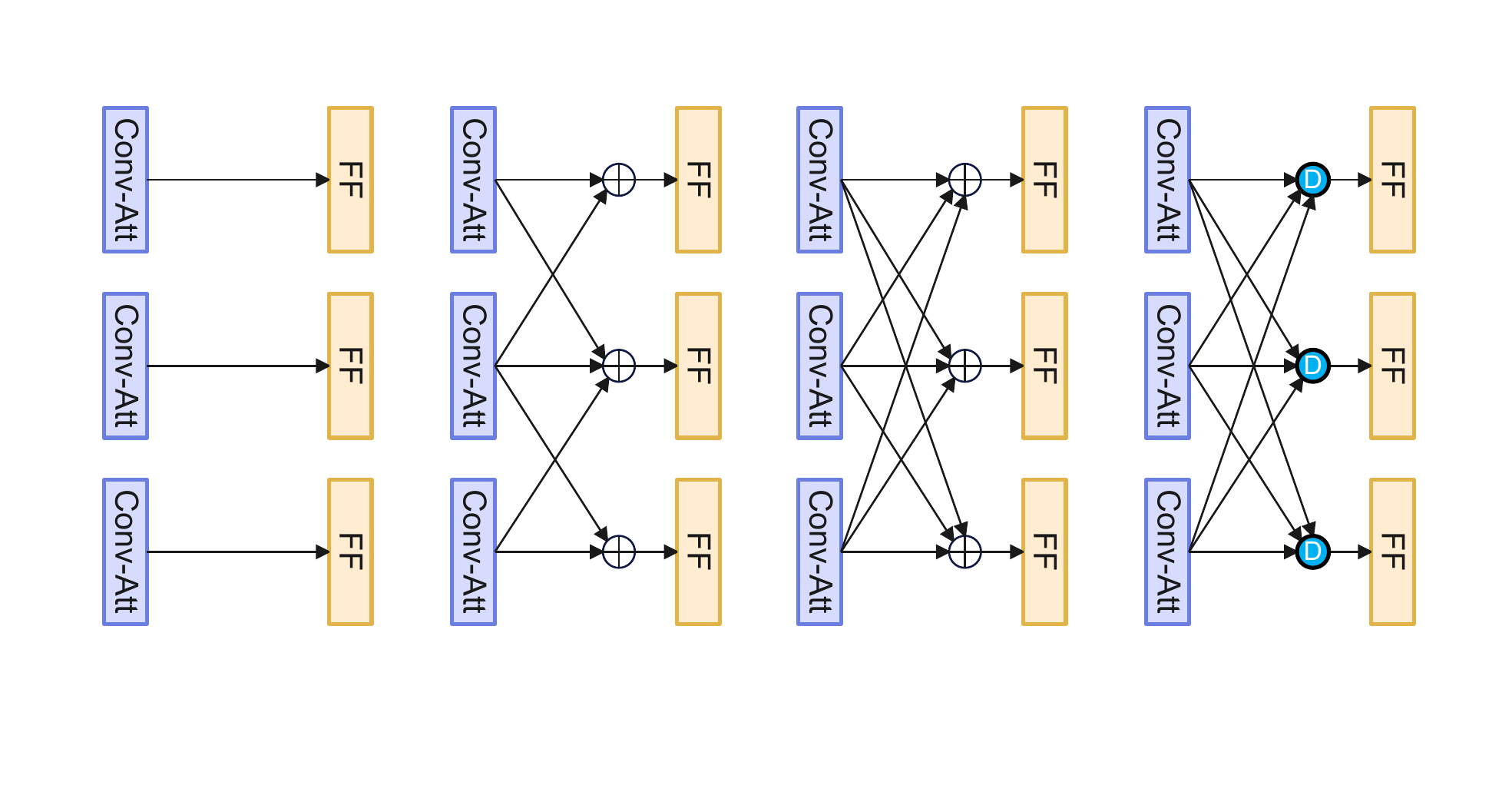}
        \put(9,6.5){\small{(a)~direct}}
        \put(28.5,6.5){\small{(b)~neighbour}}
        \put(55,6.5){\small{(c)~dense}}
        \put(75.5,6.5){\small{(d)~dynamic}}
    \end{overpic}
    \caption{Different connection manners for parallel blocks.
    (d) illustrates the proposed dynamic connection and 
    (a)-(c) are three variants of it.
    }
    \label{fig:connect}
\end{figure}

\subsection{Dynamic Connection Parallel Block}
\label{sec:dcpb}
We design the DCPB to introduce the cross-scale correspondence into the URanker.
The parallel blocks in the same group communicate with each other through the proposed dynamic connection mechanism, as shown in  Figure~\ref{fig:connect}(d).
This mechanism can adaptively integrate multi-scale features.
Different from the commonly used connection manners such as direct, neighbour, and dense as illustrated in Figure~\ref{fig:connect}(a)-(c),
this mechanism avoids the underutilization of multi-scale features and, 
conversely, prevents the pollution of the original-scale features by overusing other scale features.
More discussions about different  connection manners are provided in our ablation studies.
The details of the DCPB are described as follows.

The features of different scales are first passed through a conv-attention layer independently to keep the channel number consistent.
Then, the features from other scales are down-sampled or up-sampled by bilinear interpolation to align the feature size.
Given $\mathcal{F}_1$, $\mathcal{F}_2$, and $\mathcal{F}_O$ from different scales, 
the dynamic connection mechanism can be written as:
\begin{equation}
    \mathcal{F}_{cross} = \mathcal{F}_O + \alpha_1\mathcal{F}_1 + \alpha_2\mathcal{F}_2,
\end{equation}
where $\mathcal{F}_{cross}$ denotes the output cross-scale features, $\alpha_1$ and  $\alpha_2$ are two learnable parameters to adjust the amplitude.
At last, the $\mathcal{F}_{cross}$ is processed by a feed-forward layer and fed to the next DCPB.

\subsection{Learning to Rank}   
\label{sec:ranker-learning}
The ranker learning technique~\cite{liu2017rankiqa,zhang2019ranksrgan} is employed to optimize the URanker.
It drives the URanker to learn to rank the visual quality of the enhanced results for the same input.

During the training stage, 
an image pair $\{I_n, I_m\}$ is selected from a set of well-sorted images $I = \{I_1, I_2, \dots, I_N\}$.
The URanker predicts the correspondence scores $\{s_n, s_m\}$.
Then the scores are fed to margin-ranking loss to constrain the URanker to generate scores that match the ranking relationship.  
The margin-ranking loss can be formulated as:
\begin{equation}
    \mathcal{L}(s_n, s_m) = 
    \begin{cases}
        max(0, (s_m-s_n) + \epsilon), & q_n > q_m \\
        max(0, (s_n-s_m) + \epsilon), & q_n < q_m
    \end{cases},
\end{equation}
where $q_n$ and $q_m$ represent the visual quality of $I_n$ and $I_m$, respectively.
$\epsilon$ is a parameter, which prevents the scores from getting too close.
$\epsilon$ is set to $0.5$ in our method. 

\section{Proposed UIE Network}
The more far-reaching significance of the UIQA approach is to promote the performance of UIE algorithms.
Based on this motivation, we investigate whether the proposed UIQA network, URanker, can achieve the merit when it is used as the additional supervision to train a UIE network.

\subsection{Pre-trained URanker Loss}
Since our URanker is differentiable, the pre-trained URanker can be utilized to facilitate the training of a UIE network. The pre-trained URanker loss can be expressed as:
\begin{equation}
    \mathcal{L}_{URanker}(I_e) = \text{Sigmoid}(-M_{URanker}(I_e)),
\end{equation}
where $I_e$ is the  result enhanced by a UIE network,  $M_{URanker}(\cdot)$ is pre-trained URanker.
The total loss for training a UIE network is the weighted combination of URanker loss and content loss:
\begin{equation}
    \mathcal{L}_{total}(I_e, I_{gt}) = \mathcal{L}_{content}(I_e, I_{gt}) + \lambda \mathcal{L}_{URanker}(I_e),
    \label{eq: eq_total_loss}
\end{equation}
where $I_{gt}$ represents the ground truth and 
$\mathcal{L}_{content}$ represents the pixel-wise content loss such as L1 loss, L2 loss, or perceptual loss.
$\lambda$ is a trade-off coefficient.

\begin{figure}[t]
    \includegraphics[width=\linewidth]{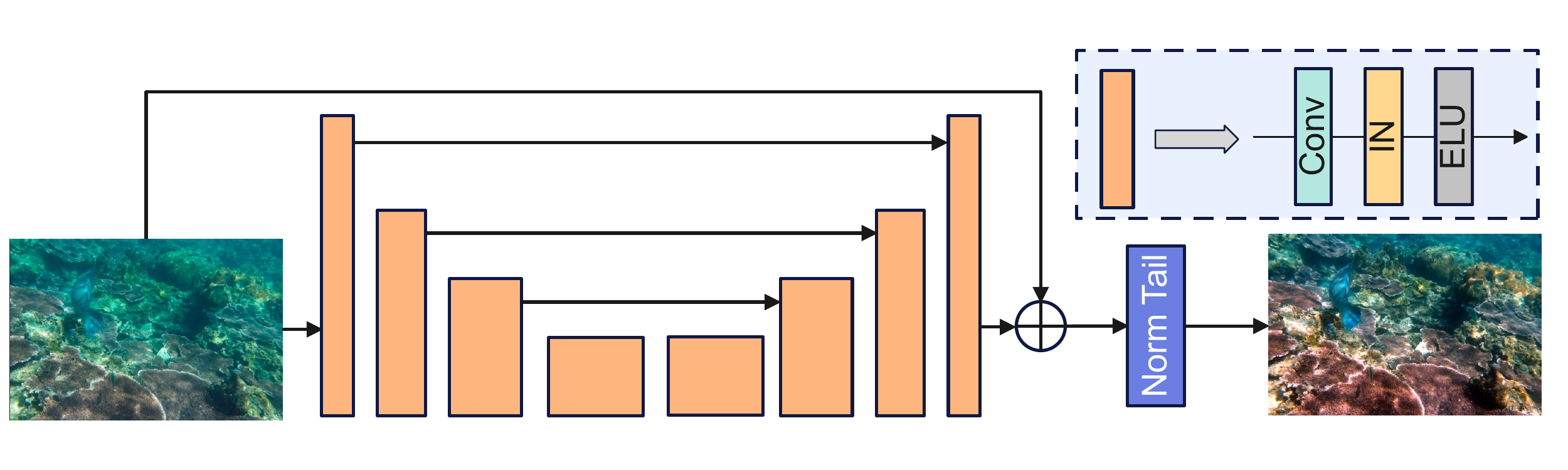}
    \caption{Overall  of the proposed NU$^{2}$Net.
    It is a residual network consisting of a stack of Conv-IN-ELU blocks.
    The normalization tail is followed at the end of the network.}
    \label{fig:uie_network}
\end{figure}

\begin{figure*}[!t]
    \centering
    \includegraphics[width=\textwidth]{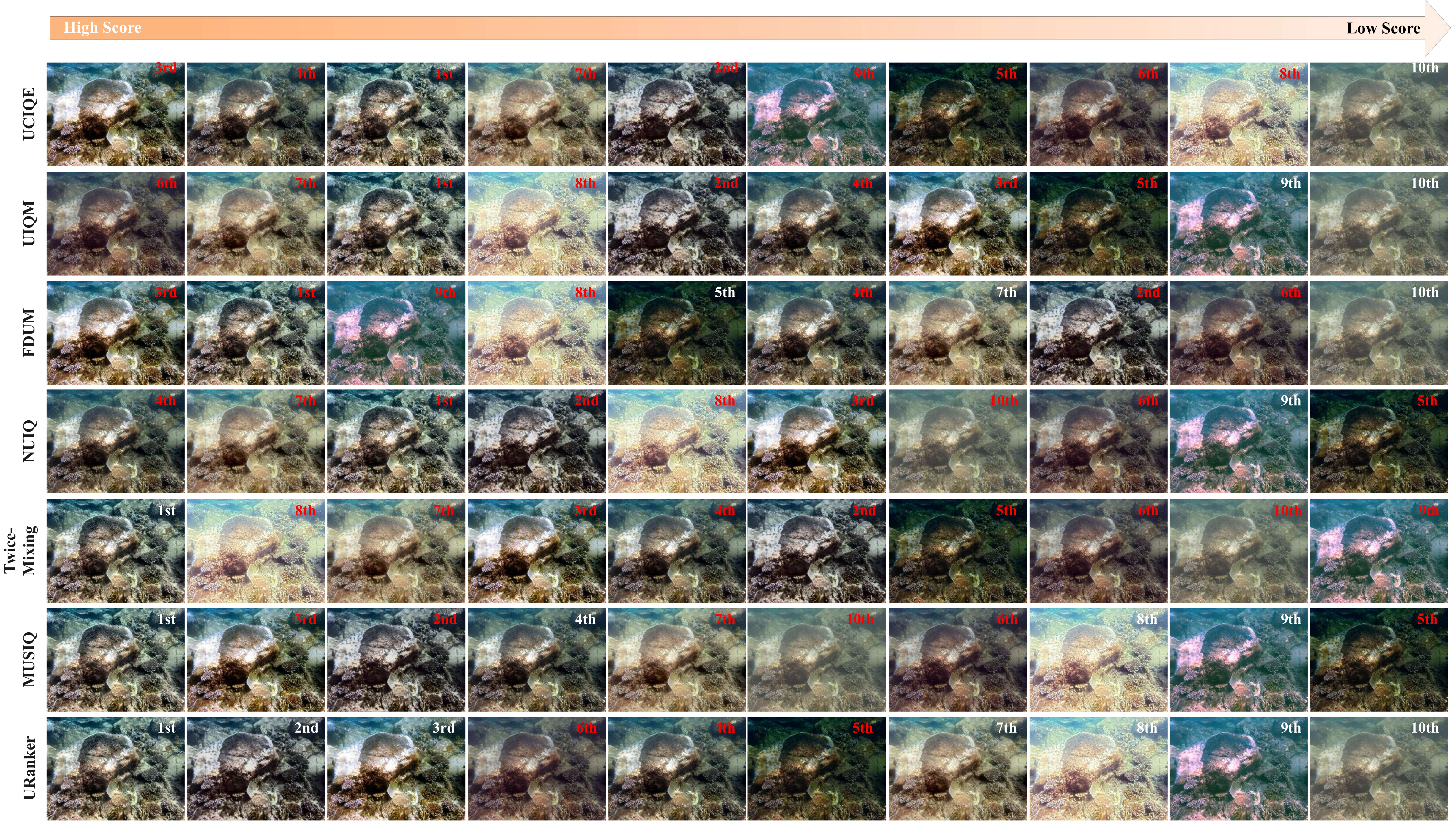}
    \caption{
        Visual comparison of UIQA.
        The prediction scores of UIQA methods decrease from left to right.
        The correspondence ground truth ranking is marked at the top-right corner of each image.
        Results with incorrect ranking are marked in red.
    }
    \label{fig:uiqa_comparison}
\end{figure*}

\subsection{UIE Network}
Our URanker loss can be easily plugged into the training phase of any UIE models.
As a baseline network, we use a simple \textbf{N}ormalization-based \textbf{U}-shape \textbf{U}IE network (NU$^{2}$Net) as shown in Figure~\ref{fig:uie_network}.
The main body of NU$^{2}$Net is a stack of Conv-IN-ELU blocks.
Besides the basic structure, we found a normalization tail can significantly improve the performance of UIE network.

To be specific, inspired by the stretching operation proposed in a traditional UIE algorithm~\cite{iqbal2010enhancing}, we formulate it as a normalization tail to handle the overflow values, \ie~outside of $[0,1]$.
When the overflow values exist in channel $c\in\{r,g,b\}$, it can be written as:
\begin{equation}
\hat{{I}_{e}^c}  = \frac{I_{e}^c - \min(I_{e}^c)}{\max(I_{e}^c) - \min(I_{e}^c)}.
\end{equation}
Please note that the normalization tail only works for the overflow values. 
%
%
More discussions are provided in the ablation studies.

\section{Experiments}
\subsection{Experimental Setting}
\subsubsection{Implementations.}
We train our URanker for $100$ epochs with the Adam optimizer with default parameters ($\beta_1=0.9, \beta_2=0.999$)
and the fixed learning rate $1.0\times 10^{-5}$.
For  data augmentation,
the input images are randomly flipped with a probability of 0.5 in both vertical and horizontal directions.
%
The proposed NU$^{2}$Net is trained for $250$ epochs with a batch size of $16$.
Adam optimizer with an initial learning rate of $0.001$ is adopted.
The learning rate is adjusted by the cosine annealing strategy.
All inputs are cropped into a size of $256\times256$ and the same data augmentation as training URanker is employed.
All experiments are implemented by PyTorch on an NVIDIA Quadro RTX 8000 GPU.

\subsubsection{Datasets \& Metrics.}
Following the experimental settings of~\cite{li2021underwater}, 
we randomly select $800$ image groups in URankerSet for training and the rest $90$ groups are regarded as the testing set.
Such settings are used for UIQE and UIE experiments.
%
To measure the consistency of the IQA results with the manual sorting results,
we adopt two rank-wise metrics, Spearman rank-order correlation coefficient (SRCC) and Kendall rank-order correlation coefficient (KRCC),  to evaluate the performance of UIQA methods, 
We discarded the linear correlation metrics generally adopted by IQA methods such as Pearson linear correlation coefficient (PLCC), as our URankerSet only provides the order without the specific scores.
For UIE experiments, the PSNR and SSIM are used as evaluation metrics.

\subsection{Underwater Image Quality Assessment Comparison}
We compare URanker with five UIQA methods, including UCIQE~\cite{yang2015underwater}, UIQM~\cite{panetta2015human}, FDUM~\cite{yang2021reference}, NUIQ~\cite{jiang2022underwater}, Twice-Mixing~\cite{fu2022twice}, and one recent Transformer-based generic IQA method, i.e., MUSIQ~\cite{ke2021musiq}.
We use the released code of UIQA methods and the PyTorch version of MUSIQ provided by IQA-PyTorch\footnote{https://github.com/chaofengc/IQA-PyTorch}.
For a fair comparison, we retrain MUSIQ using the same data as our URanker.

In Figure~\ref{fig:uiqa_comparison}, we compare the prediction results of our URanker with other methods.
We also mark the ranking order for a better visual comparison.
It can be seen that our URanker achieves the closest results to the ground truth ranking (the failure case is denoted in red) and is more consistent with the human perception as the color bar indicated.

\begin{table}[!t]
    \centering
    \begin{tabular}{p{3.3cm}p{1.2cm}<{\centering}p{1.2cm}<{\centering}}
        \toprule
        Methods & SRCC~$\uparrow$ & KRCC~$\uparrow$\\
        \midrule
        UCIQE & 0.5039& 0.4049 \\
        UIQM & 0.3705& 0.3052 \\
        FDUM & 0.3104& 0.2469 \\
        NUIQ & 0.5779& 0.4346 \\
        Twice-Mixing & 0.6218& 0.4887 \\
        MUSIQ & 0.8241& 0.6928 \\
        \midrule
        URanker (Ours) & \textbf{0.8655}& \textbf{0.7402} \\
        \bottomrule
    \end{tabular}
    \caption{Quantitative comparison of UIQA methods.  \textbf{Boldface} indicates the best result.}
    \label{table:comparison}
\end{table}

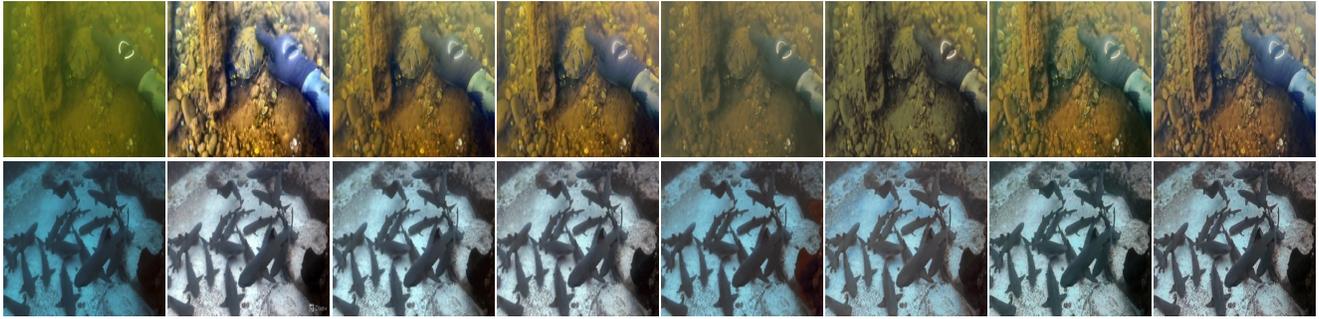
\begin{figure*}[!t]
    \begin{overpic}[width=\linewidth]{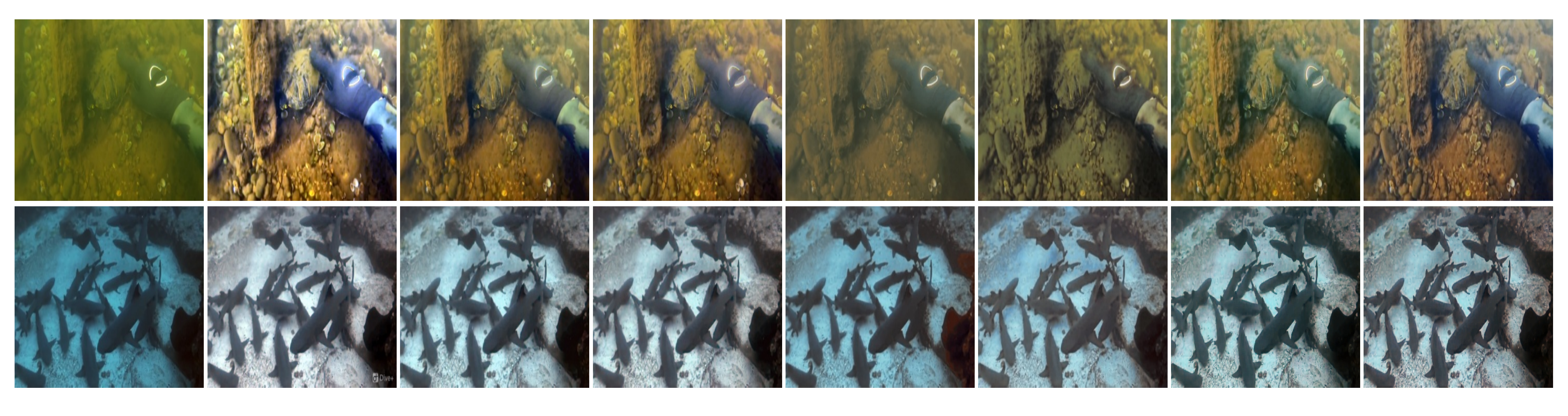}
        \put(4,-1){\small{(a)~Input}}
        \put(14.5,-1){\small{(b)~Reference}}
        \put(26.5,-1){\small{(c)~WaterNet}}
        \put(37.8,-1){\small{(d)~WaterNet+U}}
        \put(52.1,-1){\small{(e)~Ucolor}}
        \put(63.5,-1){\small{(f)~Ucolor+U}}
        \put(77.5,-1){\small{(g)~Ours}}
        \put(89,-1){\small{(h)~Ours+U}}
    \end{overpic}
    \caption{Visual comparison of UIE networks  with and without pre-trained URanker as additional supervision.
    Model+U represents the model is optimized by the pre-trained URanker loss.}
    \label{fig:uie_comparison}
\end{figure*}

Table~\ref{table:comparison} shows the quantitative comparisons of our URanker and other methods.
Our Uranker achieves state-of-the-art performance.
Compared with the second best method, MUSIQ, 
our method  brings $5.02\%$ improvements on the SRCC and $6.84\%$ improvements on the KRCC.

\begin{table}[!t]
    \centering
    \begin{tabular}{p{2.2cm}p{1.0cm}<{\centering}p{1.0cm}<{\centering}p{1.0cm}<{\centering}p{1.0cm}<{\centering}}
        \toprule
        {\multirow{2}{*}{Methods}} & \multicolumn{2}{c}{w/o $\mathcal{L}_{URanker}$} & \multicolumn{2}{c}{w/ $\mathcal{L}_{URanker}$} \\ \cmidrule(lr){2-3} \cmidrule(lr){4-5} 
        & PSNR~$\uparrow$ & SSIM~$\uparrow$ & PSNR~$\uparrow$ & SSIM~$\uparrow$\\
        \midrule
        WaterNet & 21.814 & 0.9194 & 21.926& 0.9244\\
        Ucolor & 20.433 &0.8963 & 20.686& 0.8987\\
        NU$^{2}$Net & 22.419 & 0.9227 & 22.669& 0.9246\\
        \bottomrule
    \end{tabular}
    \caption{Quantitative improvements of UIE networks by applying our pre-trained URanker as additional supervision.}
    \label{table:uie_comparison}
\end{table}

\subsection{Underwater Image Enhancement Comparison}
To further demonstrate the effectiveness of our URanker, we adopt the pre-trained URanker as additional supervision to optimize the NU$^{2}$Net and two well-known data-driven UIE networks, i.e.,  WaterNet~\cite{li2019underwater} and Ucolor~\cite{li2021underwater}.
For the content loss, WaterNet and Ucolor use the loss functions in their original works while our NU$^{2}$Net adopts L1 loss.
The trade-off coefficient $\lambda$ in Eq.~(\ref{eq: eq_total_loss}) is set to $0.025, 0.005,$ and $0.025$ for training these three UIE networks, respectively.

Figure~\ref{fig:uie_comparison} presents a set of visual examples for qualitative comparison. 
As we can see, the UIE networks with the pre-trained URanker as additional supervision  generate  results of  better visual quality.
Besides, our NU$^{2}$Net is able to generate the most visually pleasing results.

In Table~\ref{table:uie_comparison}, the pre-trained URanker as additional supervision can significantly improve the performance for all three UIE networks.
Besides, the proposed NU$^{2}$Net achieves state-of-the-art performance when compared with the WaterNet and Ucolor.

\begin{table}[t]
    \centering
    \begin{tabular}{p{3.3cm}p{1.2cm}<{\centering}p{1.2cm}<{\centering}}
        \toprule
        Baselines & SRCC~$\uparrow$ & KRCC~$\uparrow$\\
        \midrule
        w/o HPM& 0.8585& 0.7381\\ \midrule
        w/ direct & 0.8478& 0.7219\\
        w/ neighbour & 0.8589& 0.7363\\
        w/ dense& 0.8497& 0.7225\\
        \midrule
        full model & \textbf{0.8655}& \textbf{0.7402}\\
        \bottomrule
    \end{tabular}
    \caption{Ablation study on the HPM and different connection manners. \textbf{Boldface} indicates the best result.}
    \label{table:ablation}
\end{table}

\begin{figure}[!t]
    \begin{overpic}[width=\linewidth]{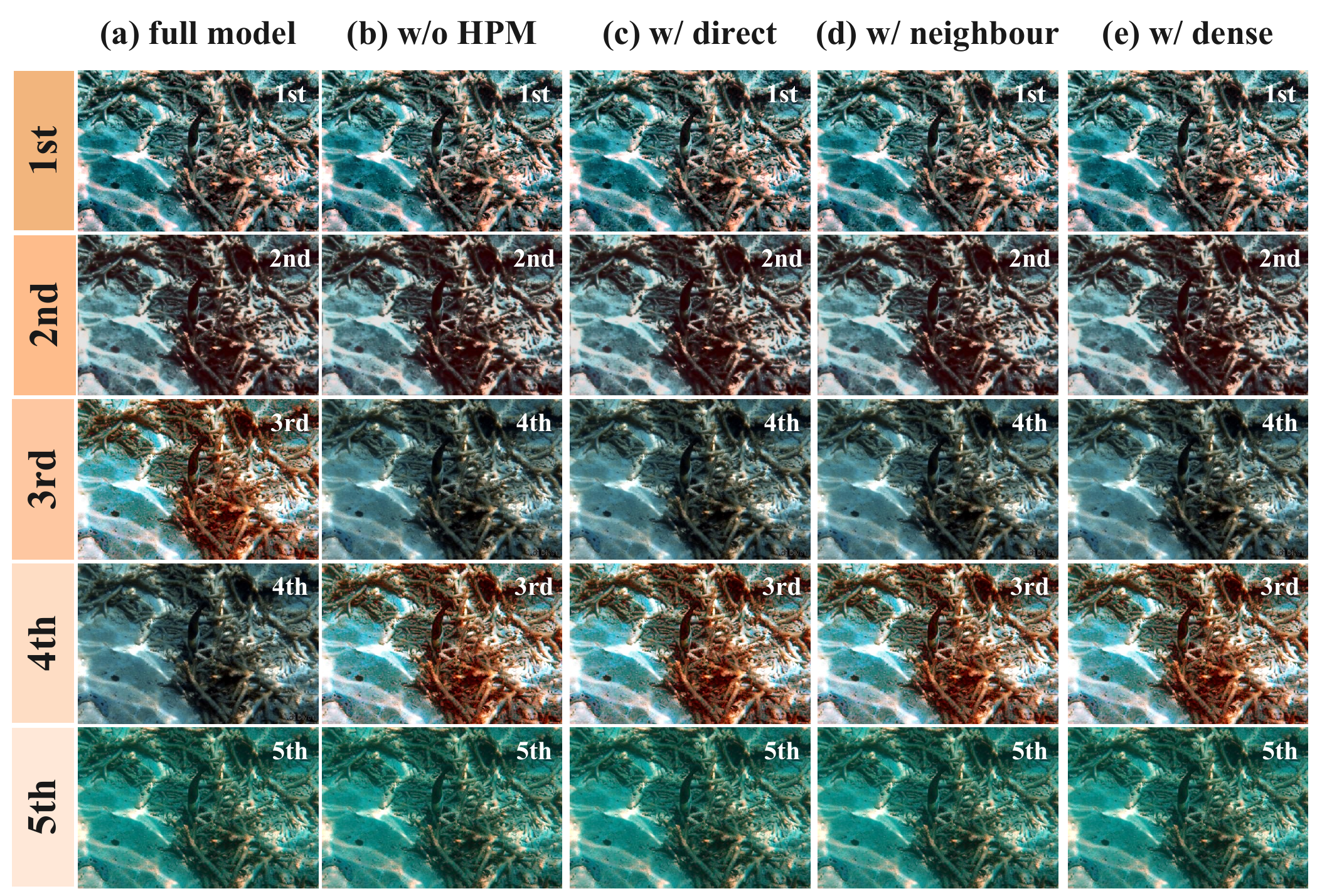}
    \end{overpic}
    \caption{Visual comparison of ablation results on HPM and DCPB.
    The prediction scores of different ablated models decrease from top to bottom.
    The ground truth ranking is marked at the top-right corner of each image.
    Results with incorrect ranking are marked in red.
    }
    \label{fig:uiqa_abalation}
\end{figure}

\begin{table}[t]
    \centering
    \begin{tabular}{p{3.3cm}p{1.2cm}<{\centering}p{1.2cm}<{\centering}}
        \toprule
        Baselines & PSNR~$\uparrow$ & SSIM~$\uparrow$\\
        \midrule
        w/o normalization tail& 18.857& 0.8682\\
        w/ Sigmoid & 18.249& 0.8292\\
        w/ clip & 19.963& 0.8837\\
        w/ IN+Sigmoid & 19.181& 0.8797\\
        w/ IN+clip & 18.257& 0.7559\\ \midrule
        full model & \textbf{22.419}& \textbf{0.9227}\\
        \bottomrule
    \end{tabular}
    \caption{Ablation study on the normalization tail and different operators. \textbf{Boldface} indicates the best result.}
    \label{table:uie_ablation}
\end{table}

\subsection{Ablation Study}
\label{sec:ablation}
We conduct a series of ablation studies to verify the effectiveness of the key components of our designs.
\subsubsection{Effectiveness of HPM \& DCPB.}
We remove the histogram prior module (HPM) or replace the dynamic connection parallel block (DCPB) with other connection manners to respectively verify their  impacts on UIQA performance.
%

In Table~\ref{table:ablation}, when the HPM is removed, 
the performance of the ablated model  decreases $0.80\%$ and $0.28\%$ in terms of SRCC and KRCC, respectively.
The full model achieves the best performance compared with the other three variants of DCPB.
In addition, the model with dense connection performs worse than the neighbour connection,
which demonstrates the over-communication of multi-scale features may pollute the original features
and further supports the necessity of dynamic connection used in our method.
In Figure \ref{fig:uiqa_abalation}, we show that only the full model can achieve accurate ranking for the five enhanced results.
The results further suggest the effectiveness of the proposed HPM and DCPB.

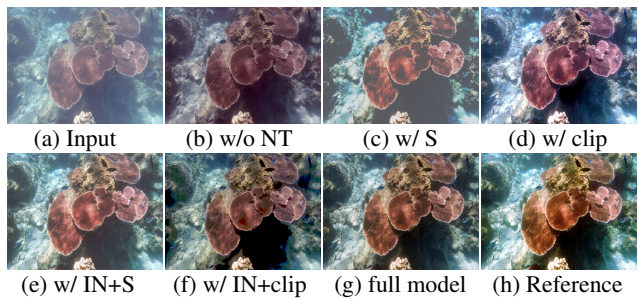
\begin{figure}[!t]
    \begin{overpic}[width=\linewidth]{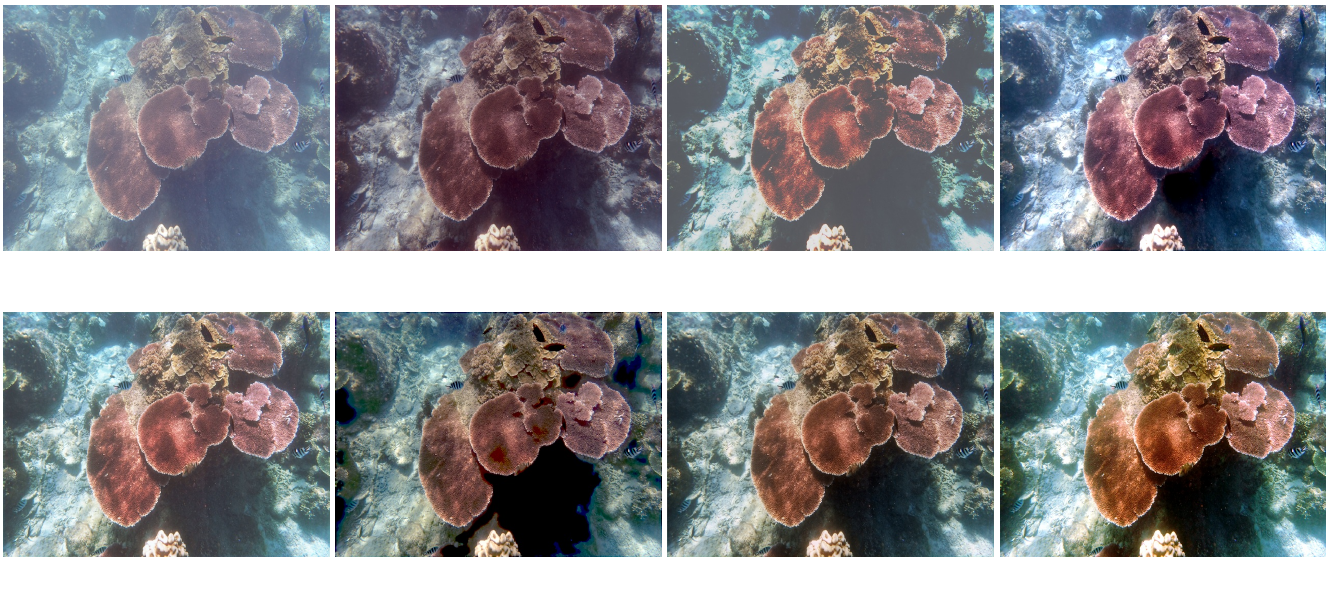}
        \put(4.5,22){\small{(a)~Input}}
        \put(28.5,22){\small{(b)~w/o NT}}
        \put(56.5,22){\small{(c)~w/ S}}
        \put(79.5,22){\small{(d)~w/ clip}}
        \put(2.5,-1){\small{(e)~w/ IN+S}}
        \put(26.5,-1){\small{(f)~w/ IN+clip}}
        \put(52,-1){\small{(g)~full model}}
        \put(77,-1){\small{(h)~Reference}}
    \end{overpic}
    \caption{Visual comparison of ablation results on the normalization tail.
    'NT' denotes the normalization tail and 'S' the Sigmoid function.
    }
    \label{fig:uie_ablation}
\end{figure}

\subsubsection{Effectiveness of Normalization Tail.}
We further analyze the effectiveness of the normalization tail used in our NU$^{2}$Net.
Firstly, the normalization tail is removed as a comparison.
Second, we replace it with $[0,1]$ clip and Sigmoid activation function to handle the outliers.
Moreover, since the stretching operation is similar to the Instance Normalization (IN)~\cite{instance2016norm}, which adjusts the distribution of data by the mean and variance,
 IN+Sigmoid and IN+clip are also used in ablation studies.

The comparison results are presented in Table~\ref{table:uie_ablation}.
The performance can increase $2.456$dB and $0.039$ at least  in terms of PSNR and SSIM by our normalization tail.
As Figure~\ref{fig:uie_ablation} shows, the full model produce the perceptually best result compared to others.
The results demonstrate this simple operation can boost performance without introducing additional parameters.

\section{Conclusion}
In this paper, we propose a new UIQA method to overcome the limitations of previous methods. The success of our method mainly lies in the task-specific designs such as the histogram prior module and dynamic connection parallel block together with a new underwater image with ranking dataset. We also demonstrate the proposed UIQA model can facilitate the performance improvement of UIE networks. Besides, the proposed normalization tail significantly boosts the UIE performance.
Both the proposed UIQA model and UIE model outperform the state-of-the-art methods.
%

\textbf{Acknowledgements.}
This work is funded by National Key Research and Development Program of China Grant No.2018AAA0100400, China Postdoctoral Science Foundation (NO.2021M701780).
We are also sponsored by CAAI-Huawei MindSpore Open Fund and the code implemented by MindSpore framework is also provided.

\appendix

\bibliography{aaai23}

\end{document}